\documentclass[10pt,twocolumn,letterpaper]{article}

\usepackage[pagenumbers]{cvpr}      

\makeatletter
\@namedef{ver@everyshi.sty}{}
\makeatother
\usepackage{tikz}

\usepackage{graphicx}
\usepackage{amsmath}
\usepackage{amssymb}
\usepackage{booktabs}

\usepackage[pagebackref,breaklinks,colorlinks]{hyperref}

\usepackage[capitalize]{cleveref}
\crefname{section}{Sec.}{Secs.}
\Crefname{section}{Section}{Sections}
\Crefname{table}{Table}{Tables}
\crefname{table}{Tab.}{Tabs.}

\newlength\savewidth\newcommand\shline{\noalign{\global\savewidth\arrayrulewidth
  \global\arrayrulewidth 1pt}\hline\noalign{\global\arrayrulewidth\savewidth}}
\def \ours {Ours\xspace} 
\usepackage{pifont}%
\newcommand{\cmark}{\ding{51}}%
\newcommand{\xmark}{\ding{55}}%
\usepackage{multirow}

\usepackage{pgfplots}
\usepackage{tabularx}
\definecolor{mycolor}{rgb}{0, 0.4, 0.7}
\definecolor{mycitecolor}{rgb}{0, 0.4, 0.7}
\definecolor{chromeyellow}{rgb}{1.0, 0.65, 0.0}
\definecolor{darkcerulean}{rgb}{0.03, 0.27, 0.49}
\definecolor{darkorange}{rgb}{1.0, 0.55, 0.0}
\definecolor{darkmidnightblue}{rgb}{0.0, 0.2, 0.4}
\definecolor{internationalorange}{rgb}{1.0, 0.31, 0.0}
\definecolor{internationalkleinblue}{rgb}{0.0, 0.18, 0.65}
\definecolor{lightsalmon}{rgb}{1.0, 0.63, 0.48}
\definecolor{mangotango}{rgb}{1.0, 0.51, 0.26}
\definecolor{mayablue}{rgb}{0.45, 0.76, 0.98}
\definecolor{majorelleblue}{rgb}{0.38, 0.31, 0.86}
\definecolor{mediumelectricblue}{rgb}{0.01, 0.31, 0.59}
\definecolor{bananamania}{rgb}{0.98, 0.91, 0.71}
\definecolor{mossgreen}{rgb}{0.68, 0.87, 0.68}
\definecolor{mintgreen}{rgb}{0.6, 1.0, 0.6}
\definecolor{palegreen}{rgb}{0.6, 0.98, 0.6}
\definecolor{bananayellow}{rgb}{1.0, 0.88, 0.21}
\definecolor{bluebell}{rgb}{0.64, 0.64, 0.82}
\definecolor{carnationpink}{rgb}{1.0, 0.65, 0.79}

\usepackage{makecell}

\def\cvprPaperID{00} 
\def\confName{CVPR}
\def\confYear{2023}

\begin{document}

\title{D$^3$ETR: Decoder Distillation for Detection Transformer}

\author{
Xiaokang Chen$^1$\thanks{\noindent Equal Contribution. Xiaokang formalized the idea and led the project.} \quad
Jiahui Chen$^2\footnotemark[1]$ \quad
Yan Liu$^3$ \quad
Gang Zeng$^1$ \quad
 \\
School of Intelligence Science and Technology, Peking University$^1$ \\ Beihang University$^2$ \quad
Microsoft Research Asia$^3$ \\
}
\maketitle

\begin{abstract}
\textcolor{black}{While various knowledge distillation (KD) methods in CNN-based detectors show their effectiveness in improving small students, the baselines and recipes for DETR-based detectors are yet to be built. In this paper, we focus on the transformer decoder of  DETR-based detectors and explore KD methods for them. 
The outputs of the transformer decoder lie in random order, which gives no direct correspondence between the predictions of the teacher and the student, thus posing a challenge for knowledge distillation.
To this end, we propose MixMatcher to align the decoder outputs of DETR-based teachers and students, which mixes two teacher-student matching strategies, i.e., Adaptive Matching and Fixed Matching.
Specifically, Adaptive Matching applies bipartite matching to adaptively match the outputs of the teacher and the student in each decoder layer, 
while Fixed Matching fixes the correspondence between the outputs of the teacher and the student with the same object queries, with the teacher's fixed object queries fed to the decoder of the student as an auxiliary group.
 Based on MixMatcher, we build \textbf{D}ecoder \textbf{D}istillation for \textbf{DE}tection \textbf{TR}ansformer (D$^3$ETR), which distills knowledge in decoder predictions and attention maps from the teachers to students. D$^3$ETR shows superior performance on various DETR-based detectors with different backbones. For example, D$^3$ETR improves Conditional DETR-R50-C5 by $\textbf{7.8}/\textbf{2.4}$ mAP under $12/50$ epochs training settings with Conditional DETR-R101-C5 as the teacher.}

\end{abstract}

\section{Introduction}
\textcolor{black}{Knowledge distillation (KD)~\cite{GeoffreyEHinton2015DistillingTK} is proposed to transfer knowledge from a large teacher model to a small student model, improving the student's performance for free in model inference. Recently there has been steady progress in KD methods, with encouraging results in many vision tasks, including image classification~\cite{romero2014fitnets,zagoruyko2016paying,cho2019efficacy,zhang2018deep,tian2019contrastive,zhou2021rethinking,yang2021knowledge,chen2021distilling,zhao2022decoupled} and object detection~\cite{chen2017learning,zhang2020improve,hao2020labelenc,dai2021general,guo2021distilling,zhang2022lgd,yang2022focal,yang2022masked}. Although various methods have been proposed, they mainly focus on CNN-based models~\cite{simonyan15vgg,he2016deep,sandler2018mobilenetv2} and are related to model structures, especially in object detection~\cite{chen2017learning,hao2020labelenc,yang2022focal,zhang2022lgd}. Applying existing KD methods to novel detectors, such as DETR-based detectors~\cite{detr,deformable-detr,meng2021conditional,chen2022conditional,liu2022dab,li2022dn}, poses challenges and may lead to trivial improvements. This paper aims to close the gap and explores KD methods for DETR-based detectors.}

\textcolor{black}{DETR~\cite{detr} is an end-to-end detector with transformer layers~\cite{vaswani2017attention}. DETR and its variants follow a pipeline of (i) extracting image features with a backbone, (ii) modeling global context with a transformer encoder, and (iii) making predictions for objects with a transformer decoder given the image features and object queries. To build KD baselines and recipes for DETR-based detectors, we go back to basics and investigate the influences of the above components. We find that the transformer decoder plays a critical role in maintaining good performance (Figure~\ref{fig:analysis-of-detr-components}), thus taking our step to explore KD methods in the transformer decoder.}

\newcommand\distance{0.005}
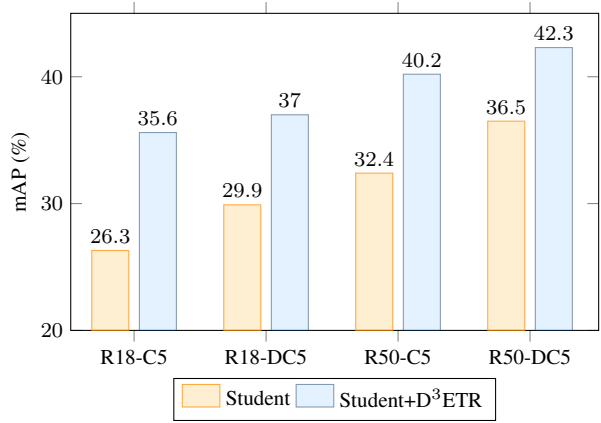
\begin{figure}
\centering

\begin{tikzpicture}[scale=1]                 
\footnotesize
\begin{axis}[
text opacity=1,
legend columns=-1,
height=5.8cm,
width=8.6cm,
legend style={at={(0.5,-0.15)},
anchor=north,font=\footnotesize},
legend cell align={left},
y label style={at={(0.08,0.5)}},
ylabel={mAP (\%)},
xtick={0.12, 0.37, 0.62, 0.87},
xticklabels={R18-C5, R18-DC5, R50-C5, R50-DC5},
xticklabel style={rotate=0,font=\footnotesize, align=center},
xmin=0,
xmax=1,
ymin=20,
ymax=45,
bar width=14pt,
ybar,
nodes near coords,
legend entries={Student, Student+D$^3$ETR},
legend image code/.code={\draw[#1] (0cm,-0.1cm) rectangle (0.5cm,0.2cm);}
]

\addplot[fill=chromeyellow, fill opacity=0.18, draw=darkorange, draw opacity=0.75, ]
coordinates
{
(0.12-\distance, 26.3)
(0.37-\distance, 29.9)
(0.62-\distance, 32.4)
(0.87-\distance, 36.5)
};

\addplot[fill=mayablue, fill opacity=0.2, draw=darkmidnightblue, draw opacity=0.5, ]
coordinates
{
(0.12+\distance, 35.6)
(0.37+\distance, 37.0)
(0.62+\distance, 40.2)
(0.87+\distance, 42.3)
};
  
\end{axis}
\end{tikzpicture}
\vspace{-2mm}
\captionsetup{font={small}}
  \caption{\textbf{Improvements over baselines.} Our D$^3$ETR obtains consistent gains over different backbones.}
\label{fig:teaser}
\end{figure}

Different from CNN-based object detectors, the outputs of the DETR decoder lie in random orders\footnote{As DETR views object detection as a set prediction problem, resulting in no direct correspondences between decoder outputs and ground-truth objects.}, resulting in no direct correspondence between the outputs of the teacher and the student. To solve this problem, we propose \textit{MixMatcher} to align the decoder outputs of the teacher and the student. MixMatcher mixes two matching strategies, \textit{Adaptive Matching} and \textit{Fixed Matching}. Adaptive matching determines teacher-student correspondence by computing the optimal bipartite matching~\cite{kuhn1955hungarian} between predictions in each decoder layer of teacher and student models. To alleviate the instability issue of bipartite matching in teacher-student adaptive matching, we also propose fixed matching. We feed the teacher's fixed object queries to the decoder of the student as an auxiliary group and apply fixed matching. It fixes the correspondence between the outputs of teacher and student models with the same object queries.

MixMatcher enables us to model the correspondence between the teacher and the student. Based on MixMatcher, we build \textit{decoder distillation for DETR-based methods (D$^3$ETR)}. Instead of only considering the predictions, we also consider the attention modules (including self-attention and cross-attention) in decoder layers when performing distillation. For the attention modules, we distill the knowledge contained in the attention maps. Extensive experiments on COCO~\cite{lin2014coco} validate the effectiveness of our D$^3$ETR. It brings significant gains to various DETR-based student models. For example, D$^3$ETR improves Conditional DETR-R50-C5~\cite{meng2021conditional} by $\textbf{7.8}/\textbf{2.4}$ mAP under $12/50$ epochs training settings with Conditional DETR-R101-C5~\cite{meng2021conditional} as the teacher.

In brief, our contributions are in three folds:
\begin{itemize}
    \item \textcolor{black}{We explore knowledge distillation for DETR-based detectors and make our attempts to solve the challenges of performing KD on the transformer decoder.}
    \item \textcolor{black}{We propose MixMatcher, which applies Adaptive Matching and Fixed Matching to model the relation between the DETR-based teacher and student. Then, we build a simple but effective distillation method D$^3$ETR.}
    \item The proposed method, D$^3$ETR, could be applied to DETR-based detectors and improve the performance significantly.
\end{itemize}

\section{Related Work}
\noindent \textbf{Knowledge distillation in object detection.}
Knowledge distillation is a method of model compression and transfer learning. It is first proposed to distill the knowledge from a large teacher model to a compact student model for the classification task~\cite{yim2017gift}. Over the years, many improved KD methods have been proposed that perform distillation over intermediate features~\cite{romero2014fitnets, tian2019contrastive}, relation representation~\cite{park2019relational, tung2019similarity}, attention~\cite{zagoruyko2016paying}, etc. Recently, some works have successfully applied KD to object detection~\cite{li2017mimicking, guo2021distilling, yang2022focal, yang2022masked}. ICD~\cite{ZhengNanning2021InstanceConditionalKD} proposes an instance-based conditional distillation framework, and it finds that initializing the student model with the teacher’s parameters will lead to faster convergence. DeFeat~\cite{guo2021distilling} decouples the foreground and background in the feature maps and distills them separately. FGD~\cite{yang2022focal} uses focal and global distillation to guide the student model, achieving remarkable results. MGD~\cite{yang2022masked} transforms the distillation into a feature generation task that uses the masked student features to generate the full teacher features. These efforts focus on the distillation on ordered outputs in CNN-based detectors.
ViDT~\cite{song2021vidt} proposes a variation of a transformer-based detector and applies KD on it, which directly performs distillation on patch tokens and detection queries between teacher and student.
{\color{black}
However, we notice that in DETRs, due to the output of the decoder being unordered, teacher queries and student queries do not have direct correspondence.
Incremental-DETR~\cite{dong2022incremental} and DETRDistill~\cite{2022DETRDISTILLAS} propose to construct the correspondence between teacher and student predictions through bipartite matching. However, they ignore the issue that the bipartite matching may be unstable in the early training stage~\cite{li2022dn}. We follow this idea and go a step further: we propose MixMatcher which helps alleviate such issues.
}

\noindent \textbf{DETR-based object detection.}
With the pioneering work DETR~\cite{detr} introducing transformers ~\cite{vaswani2017attention} to object detection, more and more follow-up works~\cite{deformable-detr, meng2021conditional, liu2022dab, li2022dn, zhang2022dino} have built various advanced extensions based on DETR because it removes the need for many hand-designed components like non-maximum suppression or initial anchor boxes generation. 
Deformable-DETR~\cite{deformable-detr} introduced the multi-scale deformable attention scheme, which attends to only a small set of points around a reference and achieves better performance than DETR. 
More works focus on improving the design of decoder in DETR\cite{chen2022conditional, wang2021anchor}. 
Conditional DETR~\cite{meng2021conditional} rebuilt positional queries based on the reference points to facilitate extreme region discrimination. DAB-DETR~\cite{liu2022dab} further extends the query as a 4D anchor box that also improves the performance. The follow-up DN-DETR~\cite{li2022dn} and DINO-DETR~\cite{zhang2022dino} introduced a novel query denoising algorithm to accelerate the training of the decoder.
{\color{black}
Group DETR~\cite{QiangChen2022GroupDF,QiangChen2022GroupDV} and H-DETR~\cite{jia2022detrs} claim the multiple positive queries are the key to the fast convergence and introduce auxiliary groups as the decoder input~\footnote{These groups share the same decoder but have no interactions, and we borrow this design in our fixed matching.}.
}
These efforts show that the decoder design is very important to DETR.
Different from the existing work on designing novel schemes in the decoder, 
we start from another orthogonal point of view and propose to transfer the knowledge in the decoder from a large model to a small model.

\begin{figure*}[t]
\centering
\includegraphics[width=0.9\linewidth]{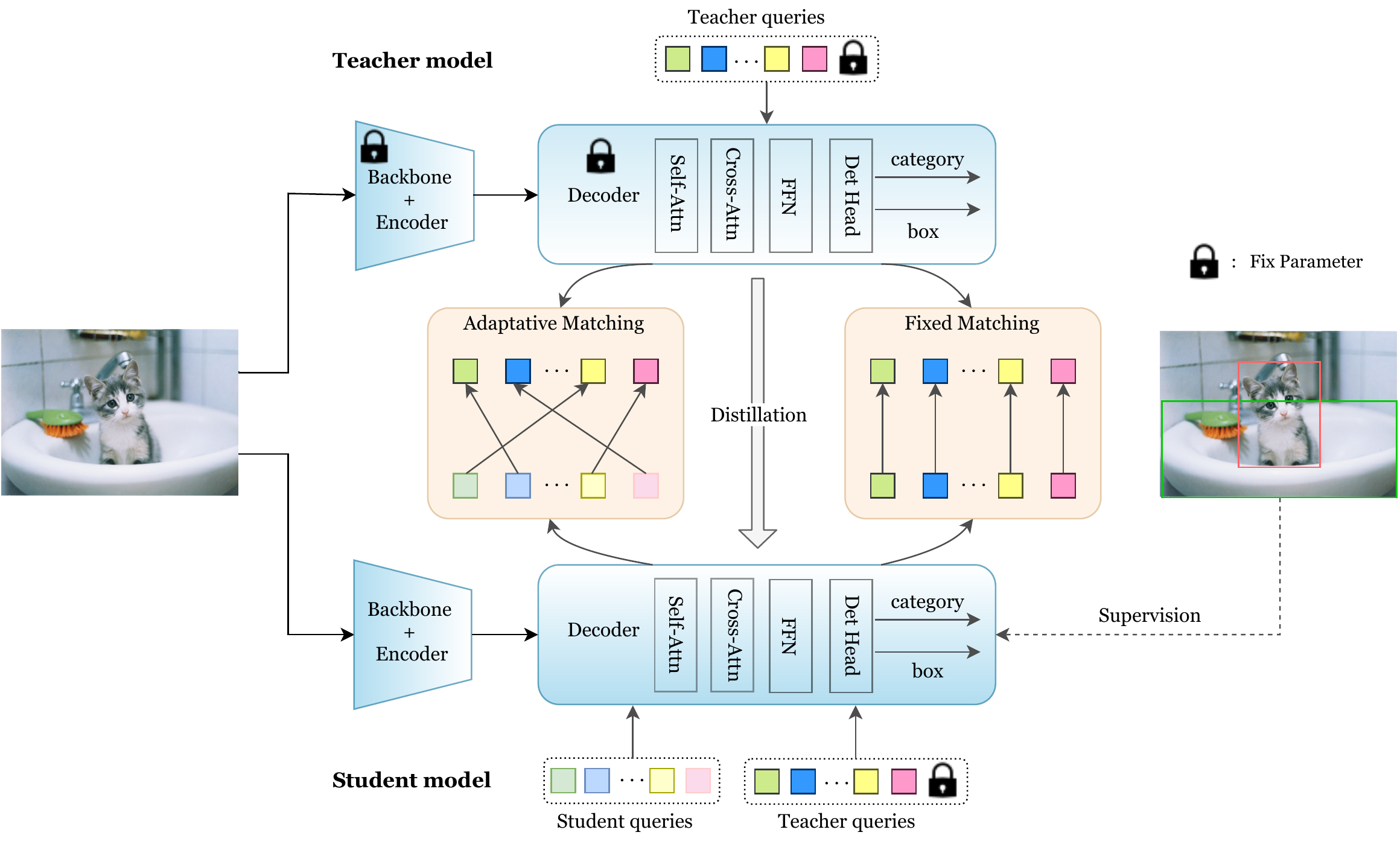}
   \caption{Architecture of the proposed method. We propose the mixed teacher-student matching strategy that composes of two components, adaptative matching and fixed matching. 
   We adopt two groups, where the first group feeds student queries to the decoder and the second feeds teacher queries to the decoder.
   Adaptative matching and fixed matching are applied to these two groups respectively.
   These two groups share the same decoder but have no interactions.
   Then we distill the knowledge in decoder self-attention, cross-attention, and prediction from the teacher model. The second group is only used in training.}
\label{fig:d3etr-arch} 
\end{figure*}

\section{Preliminary}
In this section, we first review the architecture of DETR and the attention mechanism. Then we give some analysis of the DETR structure and investigate which part has the most significant impact on the performance.

\subsection{DETR Architecture}
The DETR architecture consists of a backbone ({$e.g.$}, ResNet~\cite{he2016deep}, a transformer encoder, a transformer decoder, and object class and box position predictors. The image features are extracted by the backbone and the transformer encoder layers model the global context. The transformer decoder takes $N$ object queries as input:
\begin{align} \label{eq:query}
    \mathbf{Q} = \{\mathbf{q}_1, \dots, \mathbf{q}_{N}\}
\end{align}
Each query in the decoder is responsible for predicting either a ground-truth object (with class and bounding box) or a ``no object" class. The query may come in different forms, including the high-dimensional feature vector~\cite{detr,meng2021conditional}, the anchor point coordinates~\cite{wang2021anchor}, and the box coordinates~\cite{liu2022dab}. The object queries are combined into the decoder embeddings, forming the queries of the self-attention and cross-attention layers in the decoder. The output query embeddings are fed into detection heads to produce $N$ object predictions. 

\subsection{Attention Mechanism}
The attention~\cite{VaswaniSPUJGKP17}
is computed using
the scaled dot-product. 
The inputs contain:
a set of $N_q$ queries $\mathbf{X}_q \in \mathbb{R}^{d \times N_q}$,
a set of $N_{kv}$ keys $\mathbf{X}_k \in \mathbb{R}^{d \times N_{kv}}$,
and a set of $N_{kv}$ values $\mathbf{X}_v \in \mathbb{R}^{d \times N_{kv}}$.
The attention weights
are computed based on 
the softmax of dot-products between queries and keys:
\begin{align}
    a_{ij} = \frac{e^{\frac{1}{\sqrt{d}}\mathbf{x}_{qi}^\top \mathbf{x}_{kj}}}{Z_i}~~
    \text{where~} Z_i = \sum\nolimits_{j=1}^{N_{kv}} e^{\frac{1}{\sqrt{d}}\mathbf{x}_{qi}^\top \mathbf{x}_{kj}},
\label{eq:attn_weight}
\end{align}
where $i$ is a query index
and $j$ is a key index.
The attention output 
for each query $\mathbf{x}_{qi}$ is the aggregation 
of values weighted by attention weights:
\begin{align}
    \operatorname{Attn}(\mathbf{x}_{qi}, 
    \mathbf{X}_k,
    \mathbf{X}_v)
    = \sum\nolimits_{j=1}^{N_{kv}} \alpha_{ij}\mathbf{x}_{vj}.
\label{eq:attn_format}
\end{align}

The multi-head attention consists of $M$ parallel attention heads,
\begin{align}
    &\operatorname{MultiHeadAttn}(\mathbf{x}_{qi}, 
    \mathbf{X}_k,
    \mathbf{X}_v) \nonumber \\
    =~&\mathbf{W}_o\operatorname{Concat}(\operatorname{head}_1, \operatorname{head}_2,
    \dots, \operatorname{head}_M), \nonumber \\
    \operatorname{head}_m =~& \operatorname{Attn}(\mathbf{W}_m^q\mathbf{x}_{qi}, 
    \mathbf{W}_m^k\mathbf{X}_k,
    \mathbf{W}_m^v\mathbf{X}_v),
\label{eq:multi_head_attn}
\end{align}
where $\mathbf{W}_m^q,~\mathbf{W}_m^k,~\mathbf{W}_m^v \in \mathbb{R}^{d'_m\times d}$
and $\mathbf{W^o} \in \mathbf{W}^{d \times d}$
are the projection matrices.
The dimension $d'_m$ 
is usually the same for all the heads
and set to $\frac{d}{M}$\footnote{The dimensions of the query and the key 
might be different from the dimension
the value.
The attention formulation can be similarly derived.}.

In self-attention,
the keys, values, and queries are the same.
In cross-attention,
the keys and values are the same expect that
the keys might contain positional embeddings,
and the queries are different.

\begin{figure}[t]
\centering
\includegraphics[width=\linewidth]{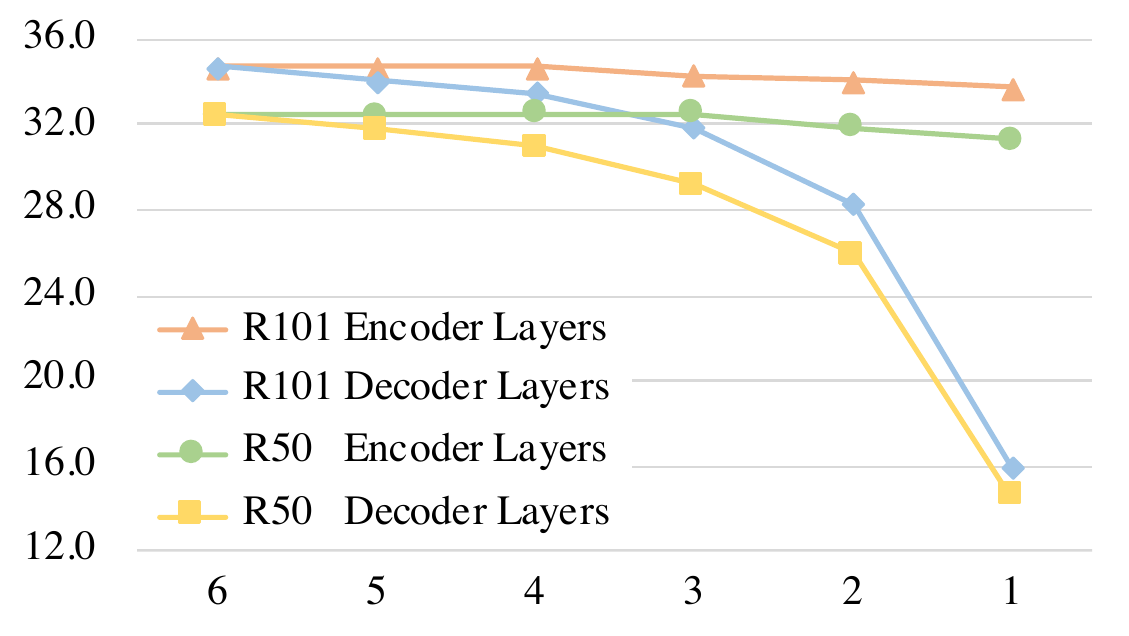}
   \caption{Analysis of different components in Conditional DETR. We adopt ResNet-$101$/ResNet-$50$ as the backbone and train the model for $12$ epochs (1$\times$ schedule). Best viewed in color.}
\label{fig:analysis-of-detr-components} 
\end{figure}

\subsection{Analysis on the DETR structure}
DETR-like methods compose of three parts: the backbone, the transformer encoder, and the transformer decoder. We conduct experiments to investigate which part has the biggest impact on the detection performance. Results are shown in Figure~\ref{fig:analysis-of-detr-components}.  We find that reducing the number of decoder layers from $6$ to $1$ leads to $17.8$/$18.3$ mAP drops in performance for R$50$/R$101$ backbones respectively. Based on this observation, we propose to distill the knowledge in the decoder.

\section{Methodology}
We first introduce MixMatcher, which consists of two teacher-student matching strategies, adaptative matching and fixed matching. Then we introduce D$^3$ETR that distills the knowledge in decoder predictions, self-attention, and cross-attention from the teacher model.

\vspace{1mm}
\subsection{MixMatcher}
\noindent \textbf{Adaptative Matching.}
The output of the DETR decoder is sparse and unordered, resulting in no direct one-to-one correspondence between teacher and student outputs. Inspired by DETR which performs bipartite matching between predicted and ground-truth objects, we propose to view the correspondence between teacher and student outputs as a bipartite matching problem.

Given the prediction $y^t$ = ($\mathbf{p}^t$, $\mathbf{b}^t$) and $y^s$ = ($\mathbf{p}^s$, $\mathbf{b}^s$) of the teacher and the student, where $\mathbf{p}$ is the soft logits for category prediction and $\mathbf{b}$ is the $4$-D vector for box prediction.
The pair-wise matching cost is defined as:
\begin{align}
\nonumber
    \mathcal{C}_{\operatorname{match}}(y_i^s, y_{\xi(i)}^t)
    = \sum\nolimits_{i=1}^{N_s}
    [\mu_{\operatorname{cls}}
    \ell_{\operatorname{bce}}(\mathbf{p}_{i}^s, \mathbf{p}_{\xi(i)}^t) + \\
    \ell_{\operatorname{box}}({\mathbf{b}}_i^s, \mathbf{b}_{	\xi(i)}^t)
    ].
    \label{eq:matching_cost}
\end{align}
Here $N_s$ is the number of student predictions. $\xi(\cdot)$ is a permutation of $N_t$ teacher predictions and usually $N_t \geqslant N_s$.
$\ell_{\operatorname{bce}}$ is the binary cross-entropy loss and $\mu_{\operatorname{cls}}=20$ is the trade-off coefficient. $\ell_{\operatorname{box}}$ is a combination of $\ell_1$ loss and GIoU loss~\cite{RezatofighiTGS019} and the loss weights are $5$ and $2$, respectively.

To find a bipartite matching between teacher and student outputs, we search for a permutation of $N_t$ elements $\hat \xi \in \Phi_{N_t}$ with the lowest cost:
\begin{align}
\hat \xi = \mathop{\operatorname{arg min}}\limits_{\xi \in \Phi_{N_t}} \sum_i^{N_s} \mathcal{C}_{\operatorname{match}}(y_i^s, y_{\xi(i)}^t)
\end{align}

To ease the training, DETR adopts the auxiliary decoding losses that each decoder layer would make detection predictions. This makes the prediction of the current stage a refinement of the previous stage. Accordingly, we adaptatively match the outputs of teacher and student models at each decoder layer.
Suppose the number of decoder layers is $L$, we could apply the adaptative matching algorithm to each decoder layer and obtain $L$ matching results: \{$\hat \xi_1, \dots \hat \xi_L$\}.

\vspace{1mm}
\noindent \textbf{Fixed Matching.}
The instability of bipartite graph matching may cause inconsistent optimization goals in early training stages~\cite{li2022dn}. To alleviate such an issue in teacher-student adaptative matching, we design an auxiliary group where we feed the fixed teacher queries into the student decoder\footnote{We assume the teacher and student models share the same query format.}. Given the same input queries, we hope the outputs of the auxiliary group and the teacher model are well aligned.

Unfortunately, there also exists instability in the bipartite graph matching between the decoder prediction and the ground truth.
This may result in such a situation: two outputs (auxiliary group and teacher model) generated from the same object query are supervised by different ground truths.
To solve this issue, 
we use the label assignment results of the teacher model to replace that of the auxiliary group in the last decoder layer:
\begin{align}
\hat \sigma^s = \hat \sigma^t,
\label{eq:fixed-matching-constraint}
\end{align}

where $\hat \sigma^t$ is the permutation of $N_t$ teacher predictions and $\hat \sigma^s$ is the permutation of $N_t$ student predictions in the auxiliary group.
Under such constraints, each query of the teacher model and the auxiliary group in the student model is supervised by the same ground truth (or ``no object"~\cite{detr}), thus strengthening the one-to-one correspondence.

We make an ingenious design to combine the two matching strategies. We feed both the student group and the auxiliary group to the student decoder during training. They share the decoder parameters, but the two groups will not interact in the decoder self-attention. 
In inference, the auxiliary group is dropped and only the student group is used.

\begin{table*}
  \centering
    \setlength{\tabcolsep}{14pt}
    \renewcommand{\arraystretch}{1.4}
    \small
  \begin{tabular}{c|cccccc}
    \shline
    Teacher & Student & Backbone & mAP & AP$_s$ & AP$_m$ & AP$_l$ \\
    \shline
    \multirow{4}{*}{\makecell{DETR \\ R$101$-C$5$ \\ $43.5$ ($500$e)}} 
      & DETR & \multirow{2}{*}{R$50$-C$5$} & $ 25.1 $ & $ 7.7 $ & $ 24.9 $ & $ 43.1 $     \\
      & + \ours &   & $30.6$ {\color{mycolor}$($+{$\mathbf{5.5}$}$)$} & $ 9.7 $ & $ 32.0 $ & $ 49.9 $     \\
      \cline{2-7}
      & DETR & \multirow{2}{*}{R$18$-C$5$} & $ 19.7 $ & $ 4.5 $ & $ 18.8 $ & $ 34.8 $     \\
      & + \ours &   & $ 28.3 $ {\color{mycolor}$($+{$\mathbf{8.6}$}$)$} & $ 7.7 $ & $ 29.3 $ & $ 48.1 $     \\
      \hline
      \multirow{4}{*}{\makecell{DETR \\ R$101$-DC$5$ \\ $44.7$ ($500$e)}}  & DETR & \multirow{2}{*}{R$50$-DC$5$} & $ 28.4 $ & $ 9.7 $ & $ 29.3 $ & $ 46.9 $     \\
      & + \ours &   & $ 39.3 $ {\color{mycolor}$($+{$\mathbf{10.9}$}$)$} & $ 17.1 $ & $ 43.1 $ & $ 59.2 $     \\
      \cline{2-7}
      & DETR & \multirow{2}{*}{R$18$-DC$5$} & $ 23.0 $ & $ 6.4 $ & $ 22.9 $ & $ 39.4 $     \\
      & + \ours &   & $ 33.0 $ {\color{mycolor}$($+{$\mathbf{10.0}$}$)$} & $ 12.4 $ & $ 34.7 $ & $ 52.8 $     \\
      \hline
    \multirow{4}{*}{\makecell{Conditional DETR \\ R$101$-C$5$ \\ $42.8$ ($50$e)}}   & Conditional DETR & \multirow{2}{*}{R$50$-C$5$} & $32.4$ & $14.7$ & $35.0$ & $48.3$     \\
      & + \ours &   & $ 40.2 $ {\color{mycolor}$($+{$\mathbf{7.8}$}$)$} & $ 19.3 $ & $ 43.5 $ & $ 59.7 $     \\
      \cline{2-7}
      & Conditional DETR & \multirow{2}{*}{R$18$-C$5$} & $ 26.3 $ & $ 10.2 $ & $ 28.2 $ & $ 39.7 $     \\
      & + \ours &   & $ 35.6 $ {\color{mycolor}$($+{$\mathbf{9.3}$}$)$} & $ 15.4 $ & $ 38.1 $ & $ 54.2 $     \\
      \hline
      \multirow{4}{*}{\makecell{Conditional DETR \\ R$101$-DC$5$ \\ $45.0$ ($50$e)}}  & Conditional DETR & \multirow{2}{*}{R$50$-DC$5$} & $ 36.5 $ & $ 17.6 $ & $ 40.0 $ & $ 52.6 $     \\
      & + \ours &   & $ 42.3 $ {\color{mycolor}$($+{$\mathbf{5.8}$}$)$} & $ 22.4 $ & $ 45.8 $ & $ 60.3 $     \\
      \cline{2-7}
      & Conditional DETR & \multirow{2}{*}{R$18$-DC$5$} & $ 29.9 $ & $ 13.4 $ & $ 32.5 $ & $ 43.3 $     \\
      & + \ours &   & $ 37.0 $ {\color{mycolor}$($+{$\mathbf{7.1}$}$)$} & $ 17.1 $ & $ 39.8 $ & $ 54.2 $     \\
    \shline
  \end{tabular}
  \caption{\textbf{Results with a $12$-epoch training schedule on MS COCO.} We highlight the improvements brought by our proposed method on two DETR-based methods. We initialize the parameters of the encoder and decoder from the teacher model.}
  \label{tab:12ep}
\end{table*}

\subsection{D$^3$ETR}
After we obtain the correspondence of the queries between the teacher and student models, we can distill the teacher's knowledge into the student model. According to the structure of the decoder, we design three distillation objectives: prediction distillation, self-attention distillation, and cross-attention distillation.

\vspace{1mm}
\noindent \textbf{Self-attention distillation.}
Decoder self-attention models the relations between object queries, which potentially plays the role of removing duplicate predictions~\cite{meng2021conditional}. Given $N$ object queries as input, we could obtain the multi-head self-attention weight map $\mathbf{A}_s^k \in \mathbb{R}^{M \times N \times N}$ of the $k$-th decoder layer, according to Eq.~\ref{eq:attn_weight} and Eq.~\ref{eq:multi_head_attn}. Similarly, we could obtain the multi-head self-attention weight map $\tilde{\mathbf{A}}_s^k$ of the teacher model. 
Please note that, although the number of teacher queries may be larger than that of the student queries, we could select queries according to the teacher-student correspondence.
Then the decoder self-attention distillation loss is defined as:
\begin{align}
    \mathcal{L}_{\operatorname{sa}} = \lambda_{\operatorname{sa}} \sum_{k=1}^{L} \texttt{MSEloss}(\mathbf{A}_s^k, \tilde{\mathbf{A}}_s^k),
\end{align}

where $L$ is the number of decoder layers. $\lambda_{\operatorname{sa}}$ is the loss weight and set as $10,000$ by default.

\vspace{1mm}
\noindent \textbf{Cross-attention distillation.}
Decoder cross-attention takes the output of the self-attention layer
as the queries
and the output of the encoder
as the keys and the values. It searches for regions of the object in the encoder output and aggregates them. Given encoder output $\mathbf{X} \in \mathbb{R}^{C \times HW}$ and $N$ queries, we could obtain the multi-head cross-attention weight map $\mathbf{A}_c^k \in \mathbb{R}^{M \times N \times HW}$ and $\tilde{\mathbf{A}}_c^k \in \mathbb{R}^{M \times N \times HW}$ of the student and teacher models, respectively.
Then the decoder cross-attention distillation loss is defined as:
\begin{align}
    \mathcal{L}_{\operatorname{ca}} = \lambda_{\operatorname{ca}} \sum_{k=1}^{L} \texttt{MSEloss}(\mathbf{A}_c^k, \tilde{\mathbf{A}}_c^k),
\end{align}
where $\lambda_{\operatorname{ca}}$ is the loss weight and set as $10,000$ by default.

\vspace{1mm}
\noindent \textbf{Prediction distillation.}
After obtaining the teacher-student correspondence, we align the student's prediction to the teacher's.
The prediction disillusion loss of the $k$-th layer is defined similarly to Eq.~\ref{eq:matching_cost}:
\begin{align}
\nonumber
\mathcal{L}_{\operatorname{pred}}^k(y_i^{sk}, y_{\xi(i)}^{tk})
    &= \sum\nolimits_{i=1}^{N_s}
    [\mu_{\operatorname{cls}}
    \ell_{\operatorname{bce}}(\mathbf{p}_{i}^{sk}, \mathbf{p}_{\xi(i)}^{tk}) + \\
    &  \quad\quad\quad\quad\quad\quad \ell_{\operatorname{box}}({\mathbf{b}}_i^{sk}, \mathbf{b}_{	\xi(i)}^{tk})
    ], \\
    \mathcal{L}_{\operatorname{pred}} &= \sum_{k=1}^{L} \mathcal{L}_{\operatorname{pred}}^k,
    \label{eq:matching-cost}
\end{align}

where $y_i^{sk}$ ($y_{\xi(i)}^{tk}$) is the $i$-th prediction of the student (teacher) in the $k$-th decoder layer.

\vspace{1mm}
\noindent \textbf{Overall distillation loss function.} The above distillation losses are applied to both the student group and the auxiliary group. The overall loss function is defined as:
\begin{align}
\nonumber
\mathcal{L}_{\operatorname{distill}} =  \mathcal{L}_{\operatorname{sa}} +  \mathcal{L}_{\operatorname{ca}} + \mathcal{L}_{\operatorname{pred}} + \\ \mathcal{L}_{\operatorname{sa}}^{\operatorname{aux}} + \mathcal{L}_{\operatorname{ca}}^{\operatorname{aux}} + \mathcal{L}_{\operatorname{pred}}^{\operatorname{aux}},
\end{align}
where $\mathcal{L}^{\operatorname{aux}}$ is the loss of the auxiliary group. 

{\color{black}
\subsection{Discussion}
DETRDistill~\cite{2022DETRDISTILLAS} is the most related work to ours. They construct the teacher-student correspondence through bipartite matching and introduce an auxiliary group to help decoder training. Ours is different from them in two aspects. (i) We propose MixMatcher which includes an auxiliary group, that helps alleviate the instability issue in teacher-student matching. In contrast, their auxiliary group have no interactions with the teacher model.
(ii) We focus on distill the knowledge in decoder attention, while they distill the knowledge in the query feature.
}

\section{Experiments}
\subsection{Setting}
\noindent\textbf{Dataset.}
We perform the experiments on the COCO $2017$~\cite{LinMBHPRDZ14}
detection dataset, which contains about $118$K training (\texttt{train}) images and
$5$K validation (\texttt{val}) images.

\noindent\textbf{Training.}
We follow the training setting of DETR~\cite{detr} and Conditional DETR~\cite{meng2021conditional} that use ImageNet pre-trained backbone
from $\operatorname{TORCHVISION}$ with Batch Normalisation (BN) layers fixed,
and the transformer parameters are initialized
using the Xavier initialization scheme~\cite{GlorotB10}.
We train the models for $12$/$50$ epochs with the AdamW~\cite{LoshchilovH17} optimizer.
The learning rate is dropped by a factor of $10$
after $11$/$40$ epochs, respectively.
We use the augmentation scheme same as DETR~\cite{detr}:
resize the input image such that
the short side is at least $480$
and at most $800$ pixels
and the long side is at most $1333$ pixels; 
randomly crop the image such that
a training image is cropped with a probability of $0.5$
to a random rectangular patch. 

\vspace{.1cm}
\noindent\textbf{Teacher models.}
For DETR~\cite{detr}, we use the officially released models trained for $500$ epochs that use ResNet-$101$-C$5$ or ResNet-$101$-DC$5$ backbone as the teacher model. For Conditional DETR~\cite{meng2021conditional}, we use the official code to train the model for $50$ epochs that use ResNet-$101$-C$5$ or ResNet-$101$-DC$5$ backbone.

\noindent\textbf{Student models.} For DETR and Conditional DETR, we train student models with the AdamW~\cite{LoshchilovH17} optimizer for $12$/$50$ epochs based on four different backbones: ResNet-$50$-C$5$, ResNet-$50$-DC$5$, ResNet-$18$-C$5$, ResNet-$18$-DC$5$.

\vspace{.1cm}
\noindent\textbf{Evaluation.}
We use the standard COCO evaluation.
We report the average precision (AP),
and the AP scores at $0.50$, $0.75$
and for the small, medium, and large objects. 

\subsection{Main Results}
Our method can be applied to various DETR-like frameworks. We first conduct experiments on two
popular detectors (DETR and Conditional DETR) with $12$-epoch 
($1\times$) and $50$-epoch 
training schedules.
~\cite{ZhengNanning2021InstanceConditionalKD} propose the inheriting strategy which initializes the student with the teacher’s neck and head parameters and improves performance. Here we use this strategy to
initialize the transformer encoder and decoder of the student with the parameter of the teacher\footnote{The inheriting strategy brings a gain of $2.1$/$0.4$ mAP on Conditional DETR-R$50$-C$5$ under $12$/$50$ epochs setting.}.

\vspace{1mm}
\noindent \textbf{Results with a standard $1\times$ schedule.}
Table~\ref{tab:12ep} reports the results. 
All the student detectors obtain significant mAP improvements with the knowledge transferred from
teacher detectors. For example, D$^3$ETR boosts detection performance when applied to Conditional DETR: +$7.8$ mAP for R$50$-C$5$, +$9.3$ mAP for R$18$-C$5$, +$5.8$ mAP for R$50$-DC$5$, and +$7.1$ mAP for R$18$-DC$5$.

\vspace{1mm}
\noindent \textbf{Results with a $50$-epoch training schedule.} We further verify the effectiveness of the proposed method under a longer training schedule. Table~\ref{tab:50ep} reports the results. We find that D$^3$ETR could still improve the baseline models significantly. D$^3$ETR improves Conditional DETR-R$50$-C$5$ by $2.4$ mAP and Conditional DETR-R$50$-DC$5$ by $1.3$ mAP, which even outperform the corresponding teacher detectors.

\begin{table*}
  \centering
    \setlength{\tabcolsep}{14pt}
    \renewcommand{\arraystretch}{1.4}
    \small
  \begin{tabular}{c|cccccc}
    \shline
    Teacher & Student & Backbone & mAP & AP$_s$ & AP$_m$ & AP$_l$ \\
    \shline
    \multirow{4}{*}{\makecell{DETR \\ R$101$-C$5$ \\ $43.5$ ($500$e)}} 
      & DETR & \multirow{2}{*}{R$50$-C$5$} & $ 34.8 $ & $ 13.9 $ & $ 37.3 $ & $ 54.5 $     \\
      & + \ours &   & $ 39.7 $ {\color{mycolor}$($+{$\mathbf{4.9}$}$)$} & $ 17.7 $ & $ 42.8 $ & $ 60.4 $     \\
      \cline{2-7}
      & DETR & \multirow{2}{*}{R$18$-C$5$} & $ 29.6 $ & $ 9.3 $ & $ 30.6 $ & $ 49.1 $     \\
      & + \ours &   & $ 33.2 $ {\color{mycolor}$($+{$\mathbf{3.6}$}$)$} & $ 11.2 $ & $ 35.2 $ & $ 54.4 $     \\
      \hline
      \multirow{4}{*}{\makecell{DETR \\ R$101$-DC$5$ \\ $44.7$ ($500$e)}}  & DETR & \multirow{2}{*}{R$50$-DC$5$} & $ 38.3 $ & $ 17.0 $ & $ 41.4 $ & $ 57.8 $     \\
      & + \ours &   & $ 42.5 $ {\color{mycolor}$($+{$\mathbf{4.2}$}$)$} & $ 21.1 $ & $ 46.6 $ & $ 61.8 $     \\
      \cline{2-7}
      & DETR & \multirow{2}{*}{R$18$-DC$5$} & $ 33.1 $ & $ 12.5 $ & $ 34.6 $ & $ 52.7 $     \\
      & + \ours &   & $ 37.3 $ {\color{mycolor}$($+{$\mathbf{4.2}$}$)$} & $ 16.0 $ & $ 40.1 $ & $ 56.8 $     \\
      \hline
\multirow{4}{*}{\makecell{Conditional DETR \\ R$101$-C$5$ \\ $42.8$ ($50$e)}} 
      & Conditional DETR & \multirow{2}{*}{R$50$-C$5$} & $ 40.9 $ & $ 20.6 $ & $ 44.3 $ & $ 59.3 $     \\
      & + \ours &   & $ 43.3 $ {\color{mycolor}$($+{$\mathbf{2.4}$}$)$} & $ 22.3 $ & $ 46.9 $ & $ 62.1 $     \\
      \cline{2-7}
      & Conditional DETR & \multirow{2}{*}{R$18$-C$5$} & $ 35.8 $ & $ 16.0 $ & $ 38.6 $ & $ 53.8 $     \\
      & + \ours &   & $ 39.6 $ {\color{mycolor}$($+{$\mathbf{3.8}$}$)$} & $ 18.8 $ & $ 42.9 $ & $ 59.2 $     \\
      \hline
    \multirow{4}{*}{\makecell{Conditional DETR \\ R$101$-DC$5$ \\ $45.0$ ($50$e)}} 
      & Conditional DETR & \multirow{2}{*}{R$50$-DC$5$} & $ 43.7 $ & $ 23.9 $ & $ 47.6 $ & $ 60.1 $     \\
      & + \ours &   & $ 45.0 $ {\color{mycolor}$($+{$\mathbf{1.3}$}$)$} & $ 25.3 $ & $ 48.7 $ & $ 63.3 $     \\
      \cline{2-7}
      & Conditional DETR & \multirow{2}{*}{R$18$-DC$5$} & $ 39.2 $ & $ 19.2 $ & $ 42.4 $ & $ 56.6 $     \\
      & + \ours &   & $ 42.0 $ {\color{mycolor}$($+{$\mathbf{2.8}$}$)$} & $ 22.3 $ & $ 45.6 $ & $ 60.4 $     \\
    \shline
  \end{tabular}
  \caption{\textbf{Results with a $50$-epoch training schedule on MS COCO.} We highlight the improvements brought by our proposed method on two DETR-based methods. We initialize the parameters of the encoder and decoder from the teacher model.}
  \label{tab:50ep}
\end{table*}

\begin{table*}[ht]
  \centering
    \setlength{\tabcolsep}{8pt}
    \renewcommand{\arraystretch}{1.25}
    \small
  \begin{tabular}{ccccccc}
    \shline
    Adaptative Matching & Prediction & Self-Attn & Cross-Attn & Fixed Matching & Inheriting & mAP \\
    \shline
     & & & & &  & $ 32.4 $ \\
     & \cmark & \cmark & \cmark & & &  $ 35.1 $ \\
    \cmark & \cmark & & & & &  $ 34.5 $ \\
    \cmark &  & \cmark & \cmark & & & $ 35.8 $ \\
    \cmark & \cmark & \cmark &  & & & $ 35.8 $ \\
    \cmark & \cmark & & \cmark & & & $ 36.0 $ \\
    \cmark & \cmark & \cmark & \cmark & & & $ 36.7 $ \\
    \cmark & \cmark & \cmark & \cmark & \cmark & & $ 38.1 $ \\
    \cmark & \cmark & \cmark & \cmark & \cmark & \cmark & $ 40.2 $ \\
    \shline
  \end{tabular}
  \caption{\textbf{Ablation study on the proposed distillation strategies.} We use Conditional DETR-R$101$-C$5$ as the teacher model and Conditional DETR-R$50$-C$5$ as the student model.}
  \label{tab:abla_disill_strategy}
\end{table*}

\begin{table}[ht]
  \centering
    \setlength{\tabcolsep}{6pt}
    \renewcommand{\arraystretch}{1.25}
    \small
  \begin{tabular}{lcc}
    \shline
    Method & \#Epochs & mAP \\
    \shline
    Conditional DETR-R$101$-C$5$ ($\star$) & $50$ & $ 42.8 $ \\
    Conditional DETR-R$50$-C$5$ ($\clubsuit$) & $12$ & $ 32.4 $    \\
    $\clubsuit$ + DeFeat & $12$ & $ 32.4 $    \\
    $\clubsuit$ + FitNet & $12$ & $ 33.3 $    \\
    $\clubsuit$ + FGD & $12$ & $ 36.0 $    \\
    $\clubsuit$ + MGD & $12$ & $ 36.7 $    \\
      \hline
    $\clubsuit$ + \ours & $12$ & $ 38.1 $    \\
    $\clubsuit$ + \ours + MGD & $12$ & $ 38.8 $    \\
    \shline
  \end{tabular}
  \caption{\textbf{Comparison with other distillation methods.} We adopt Conditional DETR-R$101$-C$5$ as the teacher model and Conditional DETR-R$50$-C$5$ as the student. Our D$^3$ETR is superior to other distillation methods, and could be further improved by combining with MGD.}
  \label{tab:abla_sota}
\end{table}

\begin{table}[ht]
  \centering
    \setlength{\tabcolsep}{8pt}
    \renewcommand{\arraystretch}{1.25}
    \small
  \begin{tabular}{ccc}
    \shline
    Adaptative Matching & Fixed Matching & mAP \\
    \shline
    &  & $ 32.4 $ \\
    \cmark &  & $ 36.7 $ \\
    & \cmark  & $ 36.3 $ \\
    \hline
    \cmark \cmark &  & $ 37.2 $ \\
    & \cmark \cmark  & $ 36.4 $ \\
    \cmark & \cmark  & $ 38.1 $ \\
    \shline
  \end{tabular}
  \caption{\textbf{Ablation study on the teacher-student matching strategy.} ``\cmark \cmark'' means we use two groups that have the same teacher-student matching strategy. The best result is obtained by using the two proposed strategies simultaneously.}
  \label{tab:abla_matching}
\end{table}

\begin{table}[ht]
  \centering
    \setlength{\tabcolsep}{8pt}
    \renewcommand{\arraystretch}{1.25}
    \small
  \begin{tabular}{lcc}
    \shline
    Method & Constraint & mAP \\
    \shline
    \hline
    Fixed Matching & \xmark & $ 37.0 $ \\
    Fixed Matching & $0$ $\sim$ $5$-th layer & $ 37.9 $ \\
    Fixed Matching & $5$-th layer & $ 38.1 $ \\
    \shline
  \end{tabular}
  \caption{\textbf{Ablation study on the constraint in fixed matching.} The best result is obtained by adding a constraint on the last decoder layer.}
  \label{tab:abla_correspondence}
\end{table}

\subsection{Ablation Study}
In this section, we first compare the proposed decoder distillation method to other CNN-based distillation methods in object detection. Then we conduct ablation studies to verify each component in our decoder distillation strategies. We adopt Conditional DETR-R$101$-C$5$ as the teacher and Conditional DETR-R$50$-C$5$ as the student. We train the student model for $12$ epochs without the inheriting strategy.

\vspace{1mm}
\noindent \textbf{The effectiveness of decoder distillation.}
We compare D$^3$ETR and other state-of-the-art object detection KD approaches in Table~\ref{tab:abla_sota}. These works all focus on the distillation of ordered outputs, and we apply them to the output feature of the transformer encoder.
We use the same teacher and student models and the same training settings in each case.
For competing distillation methods, 
we tune the hyper-parameters based on those in the corresponding papers or open-sourced code repositories  and take the best results.
From the table, we can find that the proposed method is superior to others. This illustrates that it is more efficient to perform distillation on the DETR decoder layers. Furthermore, we combine the proposed method with MGD~\cite{yang2022masked} and the performance is improved to $38.8$ mAP. This shows the possibility of further improving the performance of our method. However, it is not the focus of this paper, so we leave it as future work.

\begin{figure*}[ht]
\centering
\includegraphics[width=\linewidth]{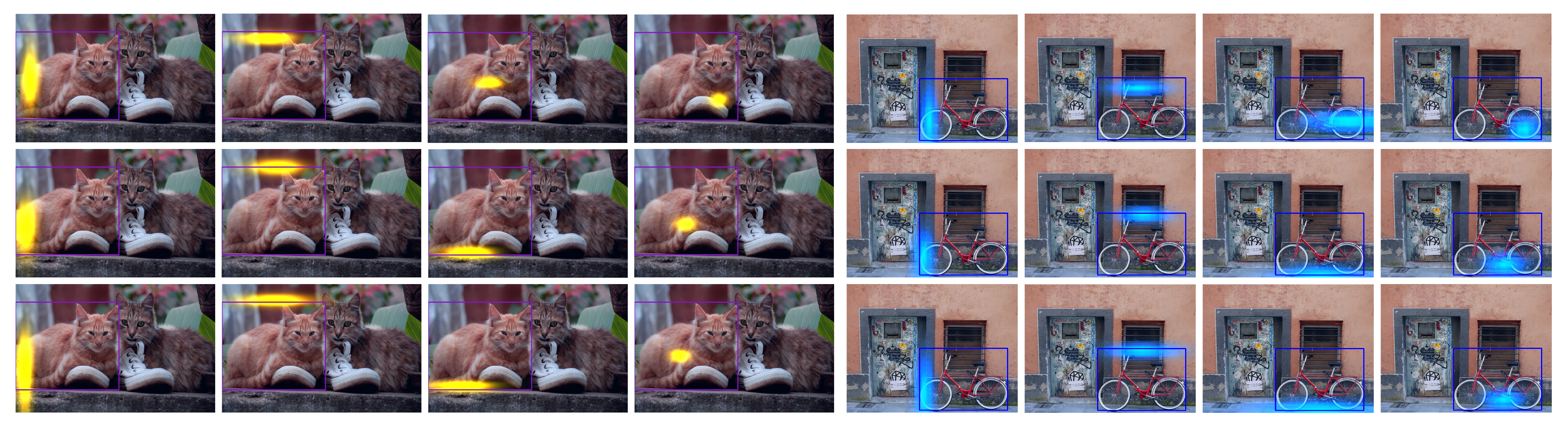}
   \caption{Comparison of spatial attention maps. The first row is the student model. The second row is the student model with our D$^3$ETR. The third row is the teacher model. We choose $4$ of $8$ heads for visualization and others are duplicates. The purple/blue boxes are ground-truth boxes. Best viewed in color.}
\label{fig:attn-vis} 
\end{figure*}

\vspace{1mm}
\noindent \textbf{The effect of each component in our method.} We gradually add the proposed strategies to the baseline and present the results in Table~\ref{tab:abla_disill_strategy}. If no teacher-student matching strategy is used, the result is $35.1$.
When adopting the adaptative matching strategy, the distillation in prediction, self-attention, and cross-attention can effectively improve the results, and the combination of the three achieves the best result ($36.7$). Based on this, fixed matching further improves the result to $38.1$. At last, the inheriting strategy brings a gain of $2.1$ and we obtain $40.2$ mAP, which is comparable to Conditional DETR-R$50$-C$5$ trained with $50$ epochs ($40.9$).

\vspace{1mm}
\noindent \textbf{Teacher-student matching strategy.} The proposed MixMatcher composes of two teacher-student matching strategies: adaptative matching and fixed matching. To validate our choice, we perform ablations on the matching strategy and report the results in Table~\ref{tab:abla_matching}.
We have some findings. First, either using adaptative matching or fixed matching could improve the baseline. Adaptative matching is slightly more efficient and achieves $36.7$ mAP. Second, when we use two groups of queries during training and adopt the same matching strategy, we find that the results can be further improved. For example, the two groups with adaptative matching are $0.5$ higher than one group. We guess this is because multiple groups of queries allow each ground truth to match more positive queries, thus easing training. Finally, using these two different strategies at the same time works best and obtains $38.1$ mAP. This illustrates the auxiliary group that uses fixed matching can help alleviate the instability issue existing in adaptative matching.

\vspace{1mm}
\noindent \textbf{Constraint in fixed matching.} We add a constraint in fixed matching to strengthen the teacher-student fixed correspondence, as illustrated in Eq.~\ref{eq:fixed-matching-constraint}. 
Without the constraint, the output of the auxiliary group may be supervised by different ground truths from the corresponding output of the teacher model. Since the two outputs are generated from the same object query, it may make the model confused. As listed in Table~\ref{tab:abla_correspondence}, the performance drops from $38.1$ to $37.0$. We also try to add constraints on all decoder layers and find the performance slightly worse than adding a constraint on the last layer.

\subsection{Visualization}
We visualize the spatial attention map~\cite{meng2021conditional} to verify whether the student learns useful information from the teacher model. The results are shown in Figure~\ref{fig:attn-vis}. 
According to ~\cite{meng2021conditional}, the spatial attention maps correspond to object extremities or a small region inside the object box. The object extremities help locate the object and the small region inside the object help recognize the category of the object.
We find that it is hard for the student to precisely locate object extremities. With our D$^3$ETR, the knowledge in the teacher model is well transferred to the student model. The student model learns similar patterns to the teacher model, thus improving the detection performance.

\section{Conclusion}

In this paper, we explore knowledge distillation for DETR-based detectors. We propose MixMatcher which models the correspondence
between the DETR-based teacher and student. Based on MixMatcher, we propose a simple but effective distillation method D$^3$ETR  and demonstrate its effectiveness with extensive experiments.

\vspace{1mm}
\noindent \textbf{Acknowledgements.} We would like to acknowledge Qiang Chen for the helpful discussions.

{\small
\bibliographystyle{ieee_fullname}
\bibliography{egbib}
}

\end{document}